\documentclass[10pt,twocolumn,letterpaper]{article}

\usepackage[pagenumbers]{cvpr} 

\usepackage{graphicx}
\usepackage{amsmath}
\usepackage{amssymb}
\usepackage{booktabs}
\usepackage{multirow}
\usepackage{xcolor}
\usepackage{url}
\usepackage{enumitem}

\usepackage{tikz}
\usetikzlibrary{positioning,arrows.meta,fit,backgrounds,calc}

\graphicspath{{./}{figs/}{Figures/}{../}{../figs/}}

\definecolor{cvprblue}{rgb}{0.21,0.49,0.74}
\usepackage[pagebackref,breaklinks,colorlinks,allcolors=cvprblue]{hyperref}

\def\paperID{39783}
\def\confName{CVPR}
\def\confYear{2027}

\newcommand{\proto}{\mathbf{p}}          
\newcommand{\member}{\mathbf{m}}         
\newcommand{\tok}{\mathbf{z}}            
\newcommand{\assign}{\mathbf{T}}         
\newcommand{\nmem}{K}                    
\newcommand{\ntok}{N}                    
\newcommand{\temp}{\tau}                 
\newcommand{\gate}{g}                    
\newcommand{\optout}{\alpha}             
\newcommand{\gtemp}{\tau_{\gate}}        

\title{Part-Set Prototypes for Explainable Mountain Gazelles Sex Classification}

\author{
Tali Boneh Shitrit \quad Amir Kedem \quad Ilan Shimshoni \quad 
\begin{tabular}[t]{@{}c@{}} 
Anna Zamansky\\{\tt\small annazam@is.haifa.ac.il}
\end{tabular}\\
University Of Haifa, Israel
}

\begin{document}
\maketitle

\begin{abstract}
Self-explainable visual models for fine-grained recognition are often hindered by rigid geometric constraints that cannot capture diverse semantic shapes in unconstrained "in-the-wild" imagery. Real-world camera-trap data introduces  challenges, including non-optimal poses, lighting variations, occlusions, animal camouflage, and environmental shortcuts. To address this, we introduce set prototypes, a general architecture where both part cardinality and unconstrained spatial layouts are mined directly from data. Our joint-matching formulation places parts simultaneously on a frozen DINOv3 token grid via truncated Sinkhorn competition, allowing unsupported parts to abstain without spatial collapse. We evaluate our approach on Mountain Gazelle sex classification, a  camera-trap benchmark, where our model achieves 0.878 macro-F1—outperforming prior  architectures and at par with an unconstrained linear probe. The strongest prototypes localize to distinctive biological areas such as horns, neck and body  allowing ecologists to visually verify decisions before trusting the metric. this formulation can generalizes naturally to diverse ecological monitoring challenges beyond a single species.\end{abstract}

\section{Introduction}
\label{sec:intro}
\begin{figure*}[t]
  \centering
  \includegraphics[width=0.75\textwidth,height=0.35\textheight]{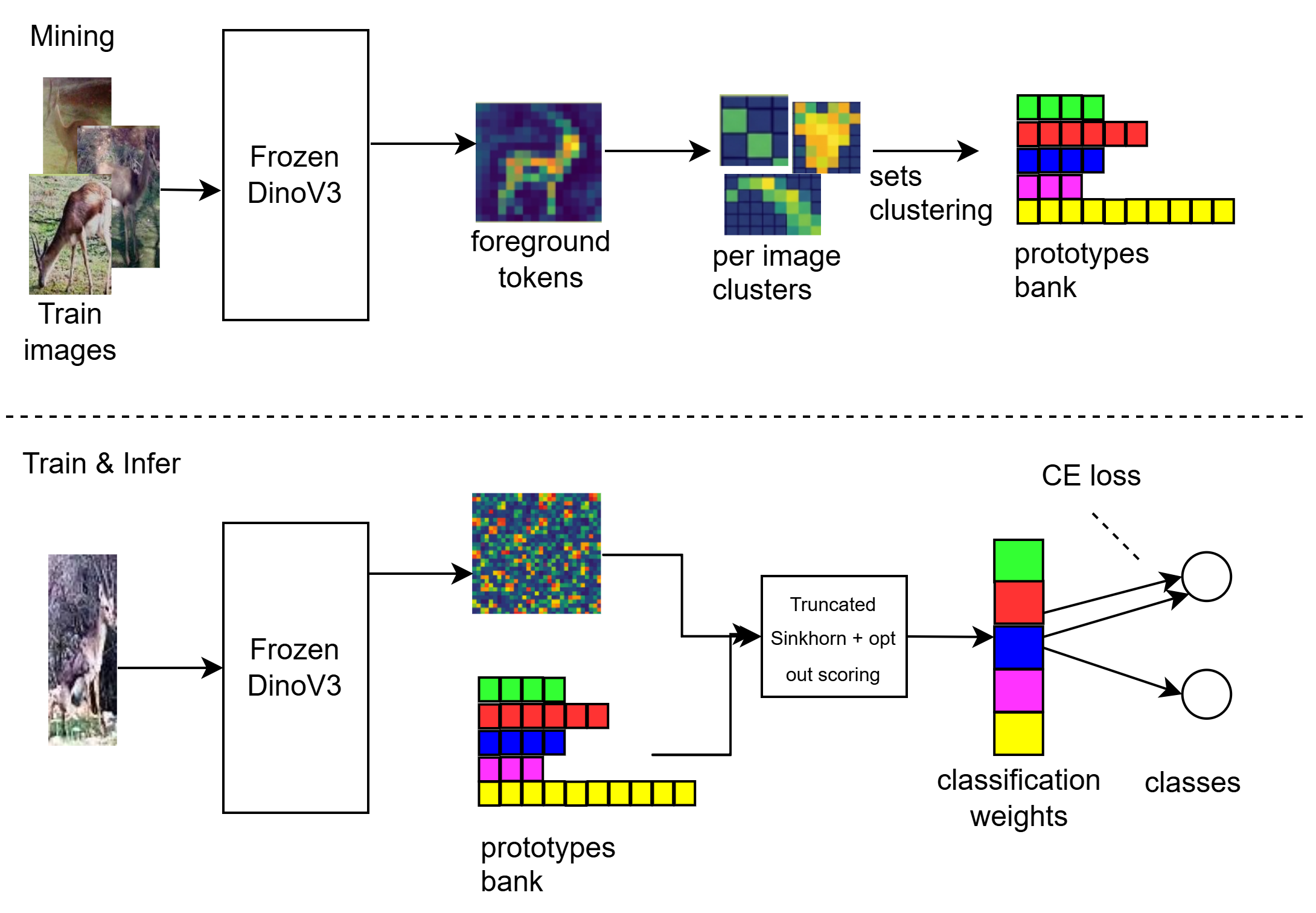}
  \caption{\textbf{Pipeline.} \emph{Initialization (top):} Foreground tokens are isolated via class-token attention from a frozen DINOv3 encoder, grouped into irregular candidate sets, and clustered into a bank of variable-cardinality prototypes initialized from real exemplars.
\emph{Train and Infer (bottom):} Member--token cosine similarities feed into a truncated Sinkhorn assignment with an opt-out gate that jointly matches each prototype's members, preventing two members from claiming the same token. Members lacking strong token support withhold mass rather than forcing a poor match, and the resulting activations pass through a non-negative, bias-free classification head.}
  \label{fig:arch}
\end{figure*}
Anthropogenic pressure has driven a global decline in biodiversity, and mitigating these trends depends on knowing not merely that a population exists but how it is structured. For an endangered ungulate such as the Mountain Gazelle (\emph{Gazella gazella})---once abundant across the Levant, now listed as Endangered and confined to fragmented, disturbed habitat~\cite{yom2021plight}---the sex ratio is a standard demographic indicator: female abundance drives herd growth, and shifts in male-to-female encounter rates signal changes in reproductive potential and herd health. Camera traps make continuous surveillance affordable, but the volume of imagery they return makes manual sorting impractical, which is why automated classification has become a cornerstone of wildlife monitoring~\cite{tuia2022perspectives, norouzzadeh2018automatically}. In this species, sex is read primarily through sexual dimorphism, which formal literature documents largely via horn morphology: adult males carry long, robust, heavily ringed, S-shaped horns (22--30~cm), whereas adult females possess markedly shorter, thinner, and smoother horns (6--11~cm) that can occasionally appear irregular or broken~\cite{castello2016bovids, baharav1974}. In unconstrained field imagery, however, horns are routinely blurred, foreshortened, or occluded. Consequently, wildlife managers rely on secondary somatic heuristics, evaluating an integrated \emph{configuration}---the overall muscular massiveness and heavy forequarters of the male versus the distinctly more slender neck, lighter profile, and finer brow-to-muzzle line of the female---rather than any single anatomical landmark.

Accuracy alone cannot tell an ecologist whether a model reads that configuration. Field imagery is not a curated benchmark: habitat, camera site, season, and time of day co-vary with where and when each sex is photographed, so a classifier can be right for the wrong reason. Post-hoc saliency is a weak remedy, because it approximates the model's reasoning after the fact, is sensitive to small perturbations, and yields inconsistent explanations across methods~\cite{GradcamSelvarajuCDVPB17, ribeiro2016should, adebayo2018sanity, kindermans2019reliability}. Self-explainable part-prototype models answer this more directly: the evidence is a part of the image that the classifier actually scored, so a domain expert can look at the evidence rather than at a reconstruction of it~\cite{li2018deep, chen2019looks, donnelly2022deformable, gautam2023looks, Nauta_2023_CVPR, ma2024interpretable, zhu2025interpretable, leago2026beyond, zimmermann2026dino}.

What such models offer the ecologist, however, is constrained by two structural rigidities: how prototypes are structured, and how they are allocated to classes. In the ProtoPNet/PIP-Net lineage, a prototype is a single patch or a fixed $k\times k$ block of adjacent patches, so its extent, layout, and part count are dictated by the architecture rather than by the animal~\cite{chen2019looks, nauta2021neural, Nauta_2023_CVPR}. Later variants relax part positions~\cite{donnelly2022deformable, zimmermann2026dino, zhu2025interpretable, leago2026beyond} or patch scale~\cite{shitrit2026prototype}, yet every prototype across the bank remains bound to a single vector or an identical, fixed part count. Capturing coherent morphological traits requires both a varying number of parts and a non-rigid spatial layout. Furthermore, existing prototype networks impose a fixed quota of prototypes per class and typically rely on backbone fine-tuning or learned feature-map adapters~\cite{zhu2025interpretable, zimmermann2026dino}---or, even when operating directly on frozen representations~\cite{leago2026beyond}, remain restricted to fixed per-class quotas of single-vector prototypes.

To address these limitations, we introduce a set-based prototype architecture that learns variable-cardinality collections of parts directly over frozen vision foundation features. Our main contributions are:

\begin{enumerate}
    \item \textbf{Variable-Cardinality Set Prototypes.} We formulate prototypes as unordered sets of member vectors whose cardinality varies across a shared bank, mined from exemplar data—operating over a global bank without rigid per-class quotas to let internal part counts adapt flexibly to the data.
    
    \item \textbf{Joint Placement by Competition.} We introduce a single-cycle truncated Sinkhorn assignment layer with an explicit opt-out mechanism, forcing member vectors to compete for distinct image tokens. This prevents multi-part collapse without auxiliary diversity losses while allowing absent parts to opt out instead of activating on background habitat.
    
    \item \textbf{Direct Prototype Optimization for Small Ecological Datasets.} By optimizing prototype sets directly on unmodified frozen foundation representations---without backbone fine-tuning or feature-map adapters---we drastically reduce trainable parameters, avoiding overfitting on scarce, unconstrained camera-trap data.
\end{enumerate}

\section{Related Work}
\label{sec:related}
\paragraph{Explainability in Automated Wildlife Monitoring.}
Deep learning has fundamentally transformed automated wildlife monitoring, particularly for species and demographic classification, counting, and individual re-identification~\cite{norouzzadeh2018automatically, tuia2022perspectives, obisa2025review}.
However, the opacity of black-box networks remains a major barrier to ecological validity. This limitation underscores the critical need for Explainable Artificial Intelligence (XAI) frameworks~\cite{minh2022explainable, schwalbe2024comprehensive, cheng2025comprehensive, csahin2025unlocking, beery2018recognition, ahmed2025evaluating, ryo2021explainable}. While inherently interpretable architectures have recently begun emerging in ecological monitoring—such as prototype-based spectrograms in bioacoustics (\eg, AudioProtoPNet~\cite{fedele2024sounds}), retrieval-augmented taxonomic reasoners~\cite{duan2026bioreason, wang2026deeptaxon}, and multi-scale part prototypes for wild Mountain Gazelle sex classification~\cite{shitrit2026prototype}—most visual interpretability in camera-trap imagery still relies predominantly on post-hoc methods. Techniques such as Grad-CAM~\cite{GradcamSelvarajuCDVPB17, pears2025explainable, nazir2024object, hasan2025transfer}, LIME~\cite{ribeiro2016should, kumar2024towards}, and pairwise feature matching~\cite{shrack2025pairwise} are widely used to verify retrospectively whether camera-trap predictions attend to anatomical regions rather than background artifacts~\cite{moallem2021explainable}. Similar post-hoc evaluations are also increasingly explored in broader animal affective and veterinary domains~\cite{feighelstein2023explainable, boneh2025segment}.

\paragraph{Prototype-Based Case Reasoning.} 
Prototype-based learning~\cite{li2018deep, chen2019looks, donnelly2022deformable, gautam2023looks, Nauta_2023_CVPR, ma2024interpretable, zhu2025interpretable} offers a transparent alternative to post-hoc explainability methods by organizing internal representations explicitly around human-understandable concepts. Rather than analyzing black-box features retrospectively, these architectures learn prototypical patterns end-to-end using only image-level labels, automatically discovering discriminative features without manual segmentation or bounding-box annotations. Self-explainable classifiers of the ProtoPNet family then enforce case-based reasoning (``this looks like that'') by comparing local image regions to these learned parts and summing the resulting evidence, making explanations faithful to the underlying computation~\cite{li2018deep, chen2019looks, rudin2019stop}. Subsequent extensions have introduced neural decision trees~\cite{nauta2021neural}, sparse scoring sheets~\cite{Nauta_2023_CVPR}, relevance propagation~\cite{gautam2023looks}, spatio-temporal modeling~\cite{wang2023learning}, adaptive ViT patch geometries~\cite{ma2024interpretable}, and multi-scale part prototypes utilizing rectangular bounding boxes for complex ecological imagery~\cite{shitrit2026prototype}. 

To better capture fine-grained semantics without heuristic constraints, recent methods integrate self-supervised vision transformers with optimal-transport and distributional principles. For instance, Zhu et al.~\cite{zhu2025interpretable} use optimal transport during prototype construction to enforce consistent token groupings and take the set mean as the prototype representative via argmax similarity, while DINO-QPM~\cite{zimmermann2026dino} and Spherical Distributional Part Prototypes~\cite{leago2026beyond} model parts through advanced distributional representations. However, these approaches typically rely on backbone fine-tuning or rigid structural assumptions that risk overfitting on data-scarce field imagery, while traditional designs remain bound by spatial rigidity and fixed class quotas. 

To resolve these limitations, we introduce set prototypes. Operating over a global prototype budget without fixed per-class quotas—following the flexible allocation of PIP-Net~\cite{Nauta_2023_CVPR} and ProtoPool~\cite{rymarczyk2022interpretable}—our formulation goes further by defining prototypes as unordered sets with variable internal cardinality ($K$). By combining data-mined set topologies with a truncated intra-image Sinkhorn layer over a completely frozen DINOv3 backbone, our approach allows part counts and spatial layouts to adapt flexibly to diverse semantic shapes without requiring auxiliary adaptation layers or risking collapse.

%
%
\section{Methods}
\label{sec:method}

\subsection{Overview}
\label{sec:overview}
Two morphological properties motivate our approach: diagnostic features (e.g., horns or facial patterns) form irregular, non-contiguous regions, and unconstrained sub-parts require joint assignment to prevent redundant matching. Prior prototype networks rely on rigid spatial grids~\cite{li2018deep, chen2019looks, donnelly2022deformable, gautam2023looks, Nauta_2023_CVPR, ma2024interpretable}, fixed-scale spatial deformations~\cite{donnelly2022deformable, zhu2025interpretable}, or rectangular multi-scale bounding boxes~\cite{shitrit2026prototype}—none of which capture arbitrarily shaped, variable-sized token sets.

Built on a frozen self-supervised ViT backbone (Sec.~\ref{sec:backbone}), our framework defines prototypes as unordered vector sets with variable cardinality (Sec.~\ref{sec:sets}). Our pipeline operates in distinct phases: first, an offline pass mines data-driven prototype topologies from foreground-isolated exemplars (Sec.~\ref{sec:mining}). During training, member vectors are jointly matched to image tokens via truncated Sinkhorn normalization with an opt-out mechanism (Sec.~\ref{sec:ot}), and class logits are accumulated through a non-negative, bias-free linear head (Sec.~\ref{sec:objective}). Test-time evaluation follows this exact formulation (Sec. 3.7), yielding faithful, natively transparent explanations directly from the learned assignments without post-hoc computation (Sec.~\ref{sec:render}).
Figure~\ref{fig:arch} illustrates our end-to-end architecture. 

\subsection{Frozen Backbone without Feature Adaptation}
\label{sec:backbone}
We encode images using a frozen DINOv3~\cite{simeoni2025dinov3} Vision Transformer to extract spatial patch tokens $\tok_n \in \mathbb{R}^{D}, n=1,\dots,\ntok$. Unlike prior work relying on encoder fine-tuning or projection heads, our prototypes reside directly in the backbone's native feature space. Bypassing feature adaptation prevents overfitting on small ecological datasets while preserving self-supervised representations. As a result, the token grid remains fixed from mining through explanation, serving as the elementary spatial unit of our framework.

\subsection{Set prototypes}
\label{sec:sets}
Unlike models that enforce fixed part counts or rigid layouts~\cite{chen2019looks, donnelly2022deformable, zhu2025interpretable, Nauta_2023_CVPR, shitrit2026prototype}, we define a prototype as an unordered set $p=\{m_1,\dots,m_K\}$ of member vectors $m_k\in\mathbb{R}^{D}$, where each member acts as a local visual primitive and the set collectively models a composite morphological concept. Operating across a shared global bank of $P$ prototypes without fixed per-class quotas, a prototype's footprint adapts freely to the data—capturing non-contiguous or arbitrarily shaped regions (e.g., curved horns or neck) because members carry no fixed spatial geometry or relative offsets. Furthermore, the internal cardinality $K$ is data-driven, determined individually per prototype during offline mining rather than imposed as a global architectural constant.

Comparing member vectors to image tokens via cosine similarity yields a pairwise score matrix $S \in [-1,1]^{K \times N}$, with entries $s_{kn}$ scoring member $m_k$ against token $z_n$. Bypassing rigid spatial grids and convolution, matching requires members to compete jointly for the tokens that best suit them (Sec.~\ref{sec:ot}).

\subsection{Prototype mining and bank initialization}
\label{sec:mining}
Prototypes are initialized from the data, in a single pass over the training set
with the frozen encoder before any gradient step. Mining happens once
and is never repeated during training.

\paragraph{Foreground by class-token attention.}
Mining is restricted to foreground tokens to focus prototype capacity on the animal rather than its background. We use the backbone's final-block class-token attention as a zero-cost saliency map, retaining a top-scoring patch fraction per image using only image-level labels. Because this separation is statistical rather than exact, background/habitat prototypes remain theoretically possible. This foreground filtering is applied exclusively during the offline mining phase. During subsequent training and inference, the complete token grid is processed.

\paragraph{Grouping foreground tokens into candidate sets.}
Within each image, retained foreground tokens are partitioned via $k$-means on their $\ell_2$-normalized embeddings, naturally clustering features by cosine similarity. To ensure candidate sets form coherent part proposals rather than noisy fragments or oversized regions, small clusters are discarded and oversized ones are truncated to tokens closest to their centroid, producing variable-sized candidate sets of arbitrary non-grid topology. This label-agnostic step proposes candidate parts purely by visual appearance, leaving part relevance and semantic utility to be learned entirely during classifier training (Sec.~\ref{sec:objective}).

\paragraph{Exemplar clustering and bank instantiation.}
Pooling candidates across the training set yields a massive, highly redundant collection. To condense this into a fixed budget of $P$ prototypes, we summarize each candidate by its mean vector and perform a second round of $k$-means clustering. We do not enforce a fixed number of prototypes per class, nor do we restrict prototypes to a single class. Because a cluster center is a single descriptor rather than a set, we instantiate each prototype from the \textit{real} candidate closest to that center—retaining its original member vectors. Consequently, prototype cardinalities are data-driven, reflecting the original mined segments subject only to pre-defined size bounds.

\subsection{Matching a set to an image}
\label{sec:ot}
Scoring each member independently at its best token ($\max_n s_{kn}$) ignores spatial competition: distinct members (e.g., left and right horns) will both collapse onto the single better-resolved horn, double-counting one piece of evidence and forfeiting the multi-part expressiveness of Sec.~\ref{sec:sets}. Members must therefore be placed \emph{jointly}, competing for image tokens.

Drawing on optimal transport principles, we formulate token allocation as a soft matching problem where member vectors distribute fractional assignment mass across the image grid. We assemble this joint placement from two main components. First, we use regularized Sinkhorn normalization~\cite{cuturi2013sinkhorn, gabriel2019computational} as a differentiable exclusivity layer rather than an iterative transport solver: alternating a row-wise softmax over tokens (temperature $\temp$) with column-wise normalization forces members to compete, scaling down tokens claimed by multiple members. We truncate this to a single sweep without enforcing strict column marginal constraints. Second, rather than appending a traditional global dustbin column to catch unmatched outliers~\cite{sarlin2020superglue}, we adapt the concept directly onto each member as an independent opt-out gate: instead of forcing a spurious token match, an absent member scales down its contribution, withholding its mass from the image entirely.

\paragraph{Letting a member decline.}
Competition alone still obliges every member to go somewhere: a gazelle in profile shows only one horn, leaving the second member with nothing to describe. Forced to allocate its mass, it lands on irrelevant background tokens, corrupting training by pulling the member vector toward habitat features.

To prevent this, member $k$ enters assignment scaled by an opt-out gate:
\begin{equation}
  \gate_k = \operatorname{sigmoid}\!\left(\frac{\max_n s_{kn}-\optout_{\proto}}{\gtemp}\right),
  \label{eq:gate}
\end{equation}
with the remaining $1-\gate_k$ mass withheld from the image entirely. This maximum is evaluated on the raw similarities \emph{before} column normalization so that the gate detects whether the feature is present.

Truncating to a single sweep makes the operation an efficient, differentiable spatial layer rather than an iterative transport solver. Each member's initial preference is divided by the total claiming mass for that token and renormalized to yield a soft assignment matrix $\assign$. This mechanism \emph{prices} overlap instead of forbidding it: a token claimed by multiple members is discounted for each, nudging them toward next-best regions. This behavior can be beneficial when resolution is limited---such as a distant gazelle whose body part occupies a single token that two members must share. The gate enters cleanly in two roles: it scales down the final assigned mass for absent parts, and because it weights the token's competing mass, a declining member exerts zero competition pressure on its siblings.

\begin{figure}[t]
  \centering
  \includegraphics[width=\linewidth]{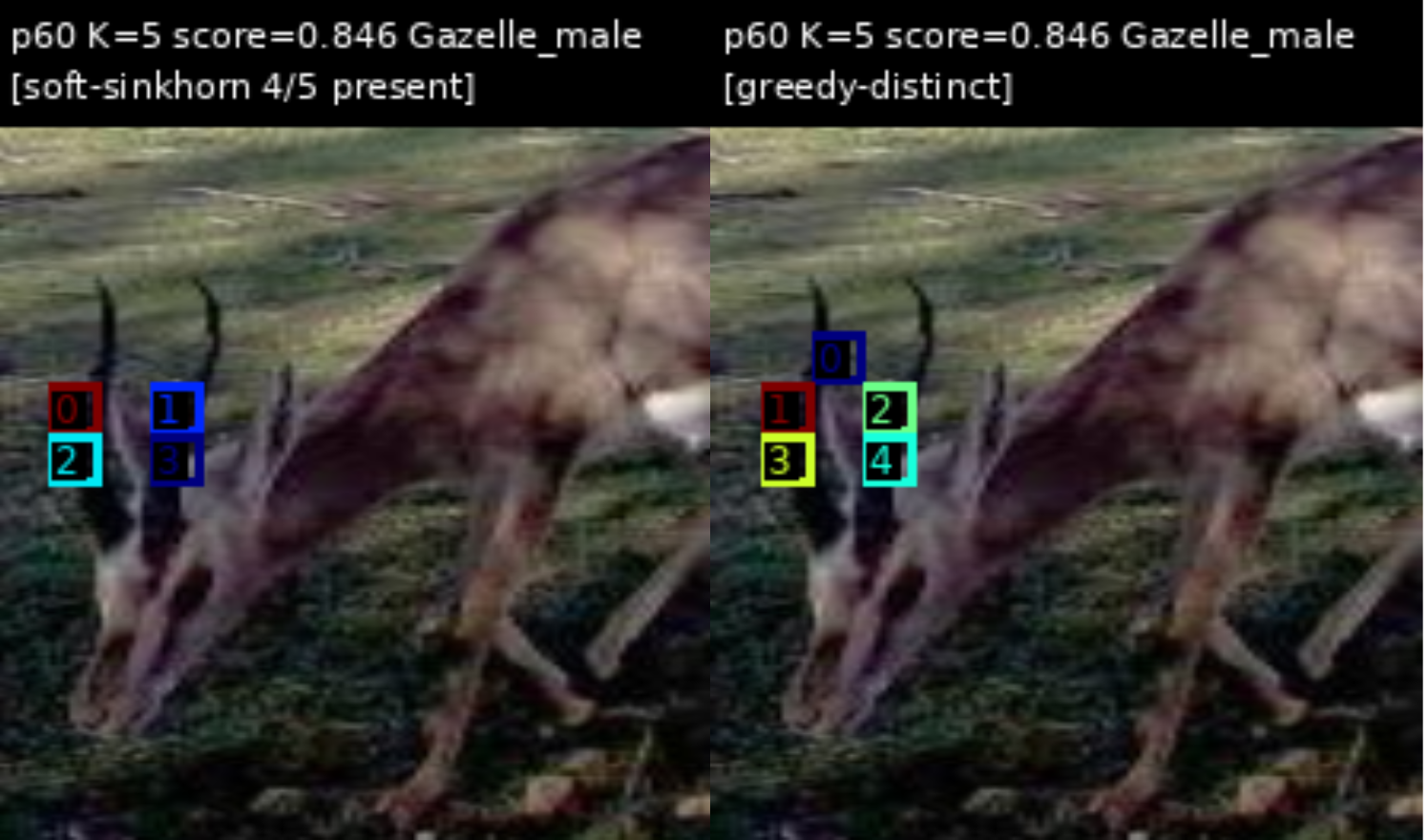}
  \caption{\textbf{Comparison of token assignment mechanisms.} Left: Our joint Sinkhorn matching with per-member opt-out gates enforces spatial competition and allows an irrelevant habitat-bound member to cleanly opt out by withholding its mass. Right: A greedy assignment baseline forces all members onto the image, leading to degenerate background matches.}
  \label{fig:assign}
\end{figure}

Each member is then scored at its assigned location, and the overall prototype activation is the mean across all $\nmem$ members:
\begin{equation}
  \hat{s}_k=\sum_{n}\assign_{kn}\,s_{kn},
  \qquad
  a(\proto)=\frac{1}{\nmem}\sum_{k}\hat{s}_{k},
  \label{eq:pooled}
\end{equation}
The average divides by the total number of members $\nmem$ regardless of whether any opted out. An absent member contributes near zero to the numerator while still counting in the denominator; abstention is thus penalized rather than free, preventing a prototype from artificially inflating its activation by dropping difficult parts.
Figure~\ref{fig:assign} visualizes the token allocation mechanics, contrasting our joint Sinkhorn assignment with per-member opt-out gating against a naive greedy baseline. Under greedy matching (right), every set member is forced to claim an image token, causing background elements—such as habitat patches—to be spuriously matched. In contrast, our exclusivity layer combined with opt-out gates (left) allows irrelevant or absent members to cleanly withhold their mass, ensuring that prototype activations rely exclusively on valid anatomical evidence rather than background shortcuts.
\subsection{Classification head and training objective}
\label{sec:objective}

Class scores are mapped from prototype activations $a(\proto_p)$ via a non-negative, bias-free linear layer:
\begin{equation}
  \ell_c = \sum_{p=1}^{P} [\mathbf{W}]^+_{cp} a(\proto_p),
  \quad [\mathbf{W}]^+_{cp} = \max(0, \mathbf{W}_{cp}).
  \label{eq:head}
\end{equation}
Because weights are non-negative and there is no bias term, each logit is an exact, additive sum of supporting prototype evidence.

With the backbone frozen and prototype cardinalities fixed, training updates only the member vectors $\member_k$, opt-out thresholds $\optout_{\proto}$, and weights $\mathbf{W}$.


\subsection{Inference}
\label{sec:inference}
At test time, the model follows the exact same formulation as training, using identical feature extraction, gating, and truncated normalization. Because the prototype assignment mechanism forms the basis of the score rather than a temporary training constraint, it remains fully intact during evaluation, with final predictions determined directly from the resulting activations.

\subsection{Rendering an explanation}
\label{sec:render}
Because a prototype is a set of learned vectors, we interpret it directly through the assignments and pooled activations used by the classifier. After training, we re-scan the training set to retrieve the top-activated images for each prototype, capturing where the learned members localize across the image grid. On these images, we visualize each member that passes the gate. To evaluate how these concepts generalize to unseen data, we rank the test set by pooled activation and overlay peak-relative heatmaps onto the top-scoring images. Presenting these top training matches alongside the test heatmaps illustrates that the prototype captures a consistent, faithful, and meaningful visual concept.

%
\section{Experimental Setup}
\label{sec:setup}
\begin{figure*}[t]
  \centering
  \includegraphics[width=0.83\textwidth,height=0.23\textheight]{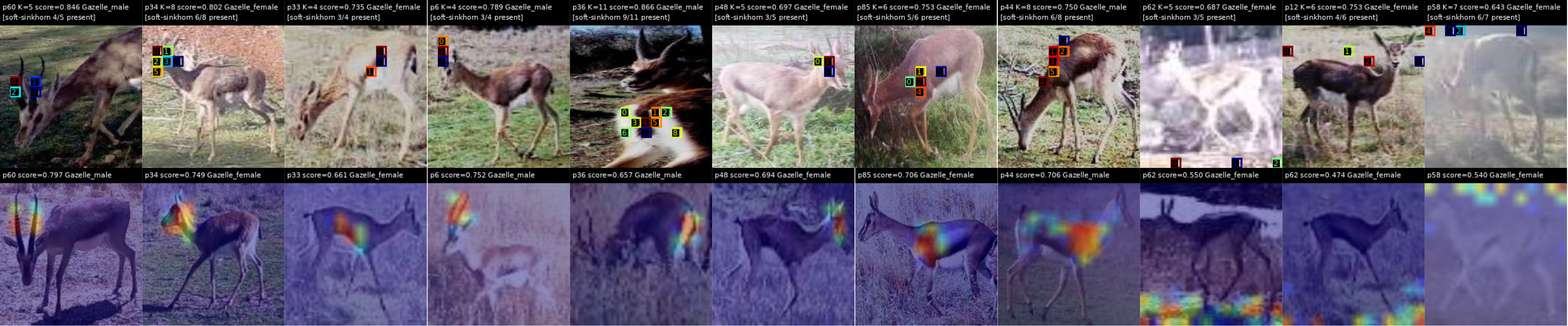}
  \caption{The most influential prototypes ordered by their contribution to the evaluation set (leftmost being the most frequently top-scoring, and rightmost the least). The top row displays each prototype as its strongest-activating training image along with its active tokens (with gated-out tokens removed). The bottom row shows representative test images and their corresponding activation maps where each respective prototype was most responsive, highlighting learned anatomical features such as horns, neck, and body structure.}
  \label{fig:protos_gallery}
\end{figure*}

\subsection{Dataset}
We use the Mountain Gazelle (\textit{Gazella gazella}) sex-classification dataset of~\cite{shitrit2026prototype}, comprising 1,833 cropped images of adult individuals (1,244 female, 589 male; fawns excluded) from 24 camera-season deployments across two campaigns (Summer 2024 and Winter 2025). Campaigns at the same physical location are treated as distinct sites. To prevent spatial data leakage, data is partitioned strictly at the camera level. We adopt the exact 8-camera test set from~\cite{shitrit2026prototype} (240 images: 117F / 123M) to maintain direct comparability with ~\cite{shitrit2026prototype} baseline. Additionally, high-volume cameras 9\textsc{w} and 10\textsc{w} are held out as our validation set for model selection (484 images: 310F / 174M). The remaining 14 cameras form the \textbf{training} set (1,109 images: 817F / 292M). Class imbalance ($\approx 70/30$) is addressed via the training loss.

\subsection{Implementation details}
\label{sec:impl}
We extract $\ntok=196$ tokens ($14\times14$, $D=1024$) from the final block of a frozen DINOv3 ViT-L/16~\cite{simeoni2025dinov3} on $224\times224$ inputs. For bank mining (Sec.~\ref{sec:mining}), we retain the top $30\%$ of tokens per image by class-token attention, run per-image $k$-means ($K \le 16$), discard clusters smaller than $4$ tokens, and truncate those larger than $12$. A second $k$-means condenses the remaining candidates into $P=96$ prototypes initialized from their nearest real exemplars, fixing $\nmem \in [4,12]$. Assignment (Sec.~\ref{sec:ot}) uses a single sweep with $\temp=0.01$, gate temperature $\gtemp=0.1$, opt-out thresholds $\optout_\proto$ initialized to $0.2$ and clamped to $[0,0.8]$.
We optimize only $\{\member_k, \optout_\proto, \mathbf{W}\}$ under class-weighted cross-entropy for $10$ epochs ($20$ epochs for the $K=1$ single-vector ablation to ensure convergence) using AdamW ($\text{lr}=8\times10^{-3}$, weight decay $1.3\times10^{-5}$) with a $2$-epoch warm-up and cosine decay. Augmentations include horizontal flips, random resized crops ($0.85$--$1.0$), mild color jitter, and small affine transforms. All hyperparameters and banks were selected on validation cameras (9\textsc{w}, 10\textsc{w}), with test cameras evaluated once on the final checkpoint.

\section{Results}
\label{sec:results}

\subsection{Detected Prototypes}
\label{sec:results:usage}
Figure \ref{fig:protos_gallery} displays the prototypes that contribute most significantly to the evaluation set. Specifically, the leftmost prototype represents the one responsible for the highest-scoring classification for the largest number of evaluation images, whereas the rightmost prototype corresponds to the least utilized (with prototypes used by fewer than 2\% of their class images being discarded). Beneath each prototype image, we show a representative test image for which that prototype was the most responsive, overlaid with its activation map.

Each prototype in the top row is visualized via the training image that activated it most strongly, highlighted with its compelling tokens. Note that tokens failing to pass the gating mechanism are omitted, and each prototype image indicates the fraction of displayed tokens relative to its full prototype set.

The recognizable groups correspond to the horns, the neck and upper chest, and the mid-body areas-- the anatomical traits listed in field guides for this species, where males exhibit large horns and more massive necks and bodies, while females carry smaller horns and a slenderer figure.

Additionally, we observe that the final three prototypes capture habitat characteristics rather than genuine discriminative features and, consequently, have the least influence. A brief retraining phase of 3 epochs excluding these prototypes yields only a negligible drop in performance (see the classification table). Because the number of decisive prototypes is relatively small, this serves as an interpretable protocol that ecologists can readily employ to refine and improve the classifier.

The prototypes exhibit strong consistency: test images consistently highlight the same anatomical body part originally captured by the prototype. This is illustrated in Figure \ref{fig:protos_gallery}, which presents 20 activation maps for each prototype, ordered identically to Figure \ref{fig:protos_gallery} (from the top-performing prototype in the largest number of images to the lowest).

\begin{figure*}[t]
  \centering
  \includegraphics[width=0.85\textwidth,height=0.45\textheight]{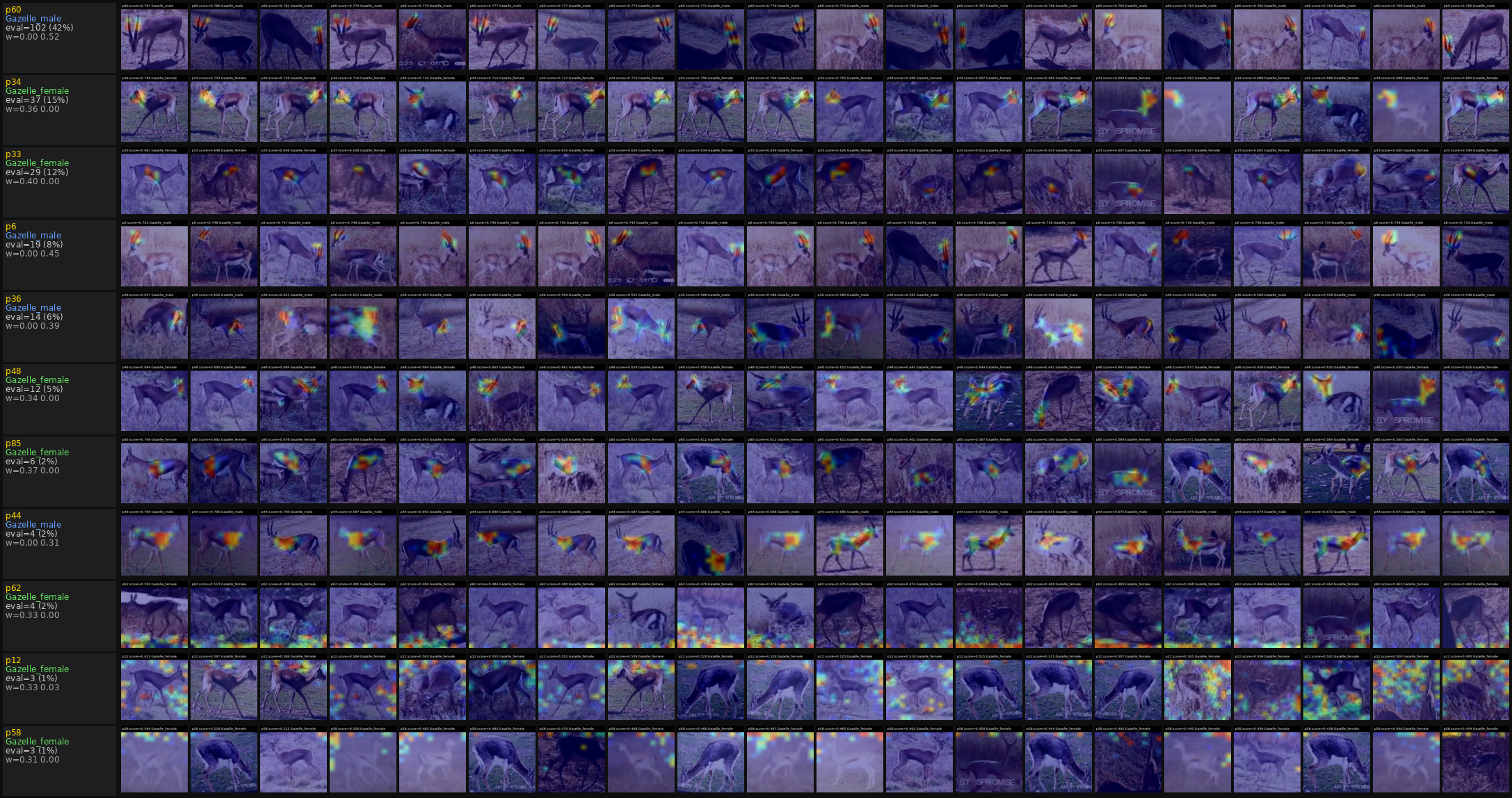}
  \caption{\textbf{Test-split prototype usage and activation maps.} Each row represents one prototype and displays its top-$20$ activating evaluation images, alongside its index, favored class, and the count (and percentage) of evaluation images where it was the strongest contributor. Per-class classification weights (female/male) are also shown, highlighting that most prototypes carry a $0.0$ weight for their unfavored class. Values above each activation map denote the prototype's activation score on that respective image. Notably, prototypes focusing on background habitat features consistently appear as the least significant contributors.}
  \label{fig:usage}
\end{figure*}

\subsection{Classification}
\label{sec:classification}

Table~\ref{tab:main} reports accuracy on the Mountain Gazelle
test split. Set prototypes reach $0.879$ accuracy and $0.878$ macro-F1,
above the results previously published on this dataset in~\cite{shitrit2026prototype}, where the best
interpretable and black-box models reach $0.754$ and $0.758$ macro-F1
respectively. On the same frozen features, a black-box linear probe
reaches $0.891$ macro-F1, so the prototype model performs on par with
an unstructured readout of the same encoder, within $1.3$ points.

To isolate the advantage of variable-cardinality sets over traditional single-vector prototypes, we evaluate a restricted baseline where set size is fixed strictly to $K=1$. Even when trained for twice as many epochs ($20$ vs.\ $10$) to reach convergence, this constrained baseline achieves only $0.838$ macro-F1—a $4.0$-point drop—confirming that multi-vector sets are essential for both capturing non-contiguous anatomical structures and optimizing efficiently. Furthermore, deleting prototypes identified as capturing background habitat and refitting only the classifier on the retained bank yields $0.874$ macro-F1, performing nearly on par with the full bank. This demonstrates that classification decisions rely on robust anatomical evidence rather than environmental shortcut learning.

\begin{table}[t]
  \centering
  \caption{\textbf{Sex classification of Mountain Gazelles on the test split.} Upper block: prior published results on this dataset. Lower block: performance using the same frozen DINOv3 features across a black-box linear probe, our single-vector ablation ($K=1$), the full set prototype model, and a habitat-pruned variant.}
\label{tab:main}
  \footnotesize
  \setlength{\tabcolsep}{3pt}
  \begin{tabular*}{\columnwidth}{@{\extracolsep{\fill}}lcccc@{}}
    \toprule
    Method & Acc. & F1$_\text{M}$ & F1$_\text{F}$ & Macro-F1 \\
    \midrule
    \multicolumn{5}{@{}l}{\emph{Prior work, as reported}} \\
    PIP-Net~\cite{Nauta_2023_CVPR} & 0.704 & 0.719 & 0.687 & 0.703 \\
    Multi-size PIP-Net~\cite{shitrit2026prototype} & 0.754 & 0.757 & 0.751 & 0.754 \\
    ConvNeXt (black box) & 0.758 & 0.768 & 0.748 & 0.758 \\
    \midrule
    \multicolumn{5}{@{}l}{\emph{This work and its controls, same frozen features}} \\
    Linear probe (black box) & 0.891 & 0.898 & 0.883 & 0.891 \\
    Set prototypes ($K=1$) & 0.841 & 0.862 & 0.813 & 0.838 \\
    Set prototypes (ours) & 0.879 & 0.889 & 0.866 & 0.878 \\
    \quad without habitat prototypes & 0.882 & 0.889 & 0.866 & 0.874 \\
    \bottomrule
  \end{tabular*}
\end{table}
\section{Discussion}

Automated wildlife monitoring has long forced ecologists into an unsatisfactory compromise: deploying opaque black-box models that achieve high statistical performance while leaving domain experts completely blind to the evidence driving their decisions, or relying on post-hoc saliency maps that offer misleading visual justifications. We set out to resolve this tension by demonstrating that ecological judgments—such as determining the sex of a Mountain Gazelle in unconstrained field photographs—can be rendered fully self-explainable without sacrificing empirical performance. Reaching an $0.878$ macro-F1 score, our set-based prototype architecture exceeds prior results on this dataset while operating within a mere $1.3$ points of an unconstrained black-box linear readout. This competitive accuracy proves that architectural interpretability does not require a performance penalty; instead, it replaces blind trust in black-box outputs with a rigorous, inspectable foundation where every unit of evidence can be examined, audited, and verified by an ecologist piece by piece.

Allowing each prototype to be a set with its own variable cardinality ($4$ to $11$ realized members) captures the physical geometry of morphological traits—such as horns, neck musculature, and body proportions—that vary in spatial extent and are non-contiguous. This advantage is empirically confirmed by our $K=1$ ablation, where restricting sets to a single vector yields a $4.0$-point drop in macro-F1 ($0.838$) and requires twice as many training epochs to reach convergence. Placing members jointly through competitive matching prevents them from collapsing onto a single salient region, achieving spatial diversity without requiring an artificial diversity loss. This keeps optimization and evaluation cleanly separable, while training over a frozen backbone keeps computational demands modest enough for a single-GPU field-ecology lab. The small $1.3$-point margin behind the unstructured probe represents the modest, honest price of enforcing an additive sum over inspectable regions, particularly since the unconstrained probe can freely exploit background regularities that our auditing protocol exposes.

Those background regularities—manifested as habitat prototypes that correlate with sex due to behavioral differences and camera placement—highlight a fundamental vulnerability of traditional black-box pipelines. In standard models, such environmental shortcuts remain hidden within the decision function, silently invalidating survey data as cameras move or vegetation shifts. A self-explainable model transforms these shortcuts from invisible failure modes into countable, actionable artifacts. Because each prototype corresponds to a localized image region, habitat-bound evidence can be explicitly identified and excised, with pruning costing only $0.4$ points of macro-F1. This ensures that demographic estimates reflect true population structures rather than the sampling artifacts of a specific camera array.

At the same time, deploying this self-explainable framework introduces specific practical boundaries. Compared to standard out-of-the-box black-box detectors, building and auditing a part-prototype bank requires substantially higher domain expertise and human-in-the-loop validation. Furthermore, the architecture fundamentally presupposes that diagnostic morphological differences actually exist in the imagery and are resolved by the frozen backbone; tasks with weak or inconsistent visual signals, such as sex classification in zebras, remain fundamentally challenging. Unlike conventional prototype-based benchmarks built on large, curated, photographer-taken datasets (such as CUB-200-2011 or Stanford Dogs), real-world ecological datasets consist of smaller collections of unconstrained camera-trap images that are frequently cropped, blurry, or captured at suboptimal angles—making a frozen self-supervised backbone not merely a computational convenience, but a critical safeguard against overfitting. Finally, while the learned bank retains near-duplicates capturing the same anatomical trait across scales or orientations (such as multiple horn prototypes), this apparent redundancy actually functions as a practical advantage in the wild: because animals frequently blend into their surroundings through camouflage and cryptic coloration, maintaining several distinct representations of key body parts helps the model reliably detect traits across shifting backgrounds.

Building on these strengths, neither the set-prototype architecture nor the auditing protocol is restricted to Mountain Gazelles. The formulation transfers naturally to other taxa and ecological monitoring challenges—including age classification, body-condition scoring, and behavioral or welfare assessment—wherever decisions depend on irregular, non-contiguous visual features in uncontrolled environments. Future work will extend this through empirical studies with working ecologists and automated habitat screening. Ultimately, by replacing opaque black-box predictions with inspectable, verifiable evidence, self-explainable models provide the foundational trust required for automated data to safely guide conservation policy.

{
    \small
    \bibliographystyle{ieeenat_fullname}
    \bibliography{main}
}

\end{document}